\documentclass[conference]{IEEEtran}
\IEEEoverridecommandlockouts
\usepackage{caption}

\usepackage{cite}
\usepackage{amsmath,amssymb,amsfonts,amsthm}
\usepackage{algorithmic}
\usepackage{graphicx}
\usepackage{svg}
\usepackage{textcomp}
\usepackage{xcolor}
\usepackage{xfrac}
\usepackage{comment}
\usepackage{booktabs} 
\usepackage{multirow} 
\usepackage{xurl}
\usepackage{bm}
\usepackage{acro}
\usepackage{tikz}
\usetikzlibrary{shapes.geometric, arrows, positioning, fit}
\usepackage[dvipsnames]{xcolor}
\usepackage{tabularx}
\usepackage{subcaption} 
\usepackage{siunitx}
\usepackage{algorithm}
\usepackage{algorithmic}
\usepackage{nicefrac}
\usepackage{bm}
\usepackage{svg}

\DeclareAcronym{FCDC}{
    short = FCDC, 
    long = Feedback Controlled Data Collection
}

\DeclareAcronym{ODD}{
    short = ODD, 
    long = operational design domain
}

\DeclareAcronym{ADS}{
    short = ADS, 
    long = automated driving system
}

\DeclareAcronym{WTTC}{
    short = WTTC, 
    long = Worst-Time-To-Collision
}

\DeclareAcronym{TTC}{
    short = TTC, 
    long = Time-To-Collision
}
 
\DeclareAcronym{HiL}{
    short = HiL, 
    long = hardware-in-the-loop
}

\DeclareAcronym{SiL}{
    short = SiL, 
    long = software-in-the-loop
}

\DeclareAcronym{MiL}{
    short = MiL, 
    long = model-in-the-loop
}

\DeclareAcronym{XiL}{
    short = XiL, 
    long = x-in-the-loop
}

\DeclareAcronym{SuT}{
    short = SuT, 
    long = system under test
}

\DeclareAcronym{PG}{
    short = PG, 
    long = Proving Ground
}

\DeclareAcronym{FOT}{
    short = FOT, 
    long = Field Operational Test
}

\DeclareAcronym{ECU}{
    short = ECU, 
    long = electronic control unit
}

\DeclareAcronym{ACC}{
    short = ACC, 
    long = adaptive cruise control
}

\DeclareAcronym{AEBS}{
    short = AEBS, 
    long = advanced emergency braking system
}

\DeclareAcronym{HAD}{
    short = HAD, 
    long = highly automated driving
}

\DeclareAcronym{ADAS}{
    short = ADAS, 
    long = advanced driver-assistance systems
}

\DeclareAcronym{ML}{
    short = ML, 
    long = machine learning
}

\DeclareAcronym{LKA}{
    short = LKA,
    long = lane keeping assist
}

\DeclareAcronym{ASIL}{
    short = ASIL,
    long = automotive safety integrity level
}

\DeclareAcronym{AV}{
    short = AV,
    long = automated vehicle
}

\DeclareAcronym{SAE}{
    short = SAE,
    long = Society of Automotive Engineers
}

\DeclareAcronym{IMU}{
    short = IMU,
    long = inertial measurement unit
}

\DeclareAcronym{ssim}{
    short = SSIM,
    long = Structural Similarity
}

\DeclareAcronym{lpips}{
    short = LPIPS,
    long = Learned Perceptual Image Patch Similarity
}

\DeclareAcronym{psnr}{
    short = PSNR,
    long = Peak Signal to Noise Ratio
}

\DeclareAcronym{mse}{
    short = MSE,
    long = Mean Squared Error
}

\DeclareAcronym{fid}{
    short = FID,
    long = Fr\'echet Inception Distance
}

\DeclareAcronym{mi}{
    short = MI,
    long = Mutual Information
}

\DeclareAcronym{ncc}{
    short = NCC,
    long = Normalized Cross-Correlation
}

\DeclareAcronym{vqgan}{
    short = VQGAN,
    long = Vector Quantized Generative Adversarial Networks
}

\usepackage[hidelinks]{hyperref}
\def\BibTeX{{\rm B\kern-.05em{\sc i\kern-.025em b}\kern-.08em
    T\kern-.1667em\lower.7ex\hbox{E}\kern-.125emX}}

\PassOptionsToPackage{svgnames}{xcolor}
\usepackage{tcolorbox}
\tcbuselibrary{skins,breakable}
\usetikzlibrary{shadings,shadows}
    {\endtcolorbox}

\DeclareRobustCommand\mytikzDotRed{\tikz[baseline=0.25ex] \fill[red] (1ex,1ex) circle (0.5ex);}
\DeclareRobustCommand\mytikzDotBlue{\tikz[baseline=0.25ex] \fill[blue] (1ex,1ex) circle (0.5ex);}

\DeclareRobustCommand\mytikzDotOrange{\tikz[baseline=0.25ex] \fill[orange] (1ex,1ex) circle (0.5ex);}

\begin{document}
\bstctlcite{BSTcontrol} 

\UseRawInputEncoding
\title{A Feedback-Control Framework for Efficient Dataset Collection from In-Vehicle Data Streams}
\author{\IEEEauthorblockN{Philipp Reis, Philipp Rigoll, Christian Steinhauser, Jacob Langner, and Eric Sax}
\IEEEauthorblockA{
FZI Research Center for Information Technology, Karlsruhe, Germany\\
Email: \{reis, philipp.rigoll, steinhauser, langner, sax\}@fzi.de}
}
\theoremstyle{definition}
\newtheorem{definition}{Definition}[section]

\maketitle
\bibliographystyle{IEEEtran}
\begin{abstract}
Modern AI systems are increasingly constrained not by model capacity but by the quality and diversity of their data. Despite growing emphasis on data-centric AI, most datasets are still gathered in an open-loop manner which accumulates redundant samples without feedback from the current coverage. This results in inefficient storage, costly labeling, and limited generalization. To address this, this paper introduces \ac{FCDC}, a paradigm that formulates data collection as a closed-loop control problem. \ac{FCDC} continuously approximates the state of the collected data distribution using an online probabilistic model and adaptively regulates which samples are retained based on feedback signals such as likelihood and Mahalanobis distance. Through this feedback mechanism, the system dynamically balances exploration and exploitation, maintains dataset diversity, and prevents redundancy from accumulating over time. In addition to demonstrating the controllability of \ac{FCDC} on a synthetic dataset that converges toward a uniform distribution under Gaussian input assumption, experiments on real data streams show that \ac{FCDC} produces more balanced datasets by $\SI{25.9}{\percent}$ while reducing data storage by $\SI{39.8}{\percent}$. These results demonstrate that data collection itself can be actively controlled, transforming collection from a passive pipeline stage into a self-regulating, feedback-driven process at the core of data-centric AI.
\end{abstract}


\section{Introduction}

The success of modern Artificial Intelligence (AI) hinges increasingly on the data it learns from rather than on the complexity of its models~\cite{yang2025qwen3technicalreport,grattafiori2024llama3herdmodels}. As architectures saturate and compute scales, progress now depends on curating datasets that are diverse, representative, and dynamically aligned with the target domain. Yet, the dominant paradigm of data collection remains fundamentally open-loop, so that vast amounts of data are gathered indiscriminately. This is especially true for vehicle data streams. Cars equipped with autonomous driving stacks generate up to $\SI{2.5}{\giga\byte\per\second}$ of data~\cite{Heinrich_24,edgar_23}. Furthermore, the phenomenon of the long-tail data distribution~\cite{Ackermann_long_tail_2017,heidecker_application-driven_2021} in vehicle experience makes the collection task even more challenging, resulting in the curse of rarity problem described in~\cite{liu_curse_2024}. Consequently, redundant and biased samples are collected, leading to unnecessary labeling and storage costs~\cite{fingscheidt_deep_2022}, and ultimately degrading dataset quality and downstream task performance~\cite{liu_curse_2024}.

While the field of data-centric AI emphasizes the importance of data quality, it still lacks a mechanism to actively control how data is collected in real-time as the model evolves. Existing adaptive data selection methods attempt to close this loop heuristically, but they lack a principled framework for convergence and optimality. What is missing is a systematic way to formalize data collection as a controlled process, one that can sense its current informational state and adjust its collection strategy accordingly. In this view, the data collection pipeline becomes a dynamical system. The state reflects the coverage of the dataset, and its control inputs govern sampling strategies to enrich the collected dataset.

Control theory provides the mathematical foundation to establish such a feedback-driven data collection paradigm. This controlled approach transforms data collection from a passive, open-loop activity into an actively managed process, which marks a paradigm shift from Big Data to Fast Data~\cite{miloslavskaya_big_2016}.

\subsection{Contribution}
This paper provides four main contributions:
\begin{enumerate}
    \item Introduction of the \textbf{\acf{FCDC} framework},  which treats data collection as a dynamic process where the evolving dataset influences the subsequent collection strategy to control dataset properties.
    \item \textbf{Formalization} of data collection as a \textbf{closed loop  process}, establishing a theoretical connection between control theory and data-centric artificial intelligence.
    \item Development of an online algorithm that integrates a \textbf{probabilistic density estimator} with an adaptive controller to regulate sampling based on real-time feedback of novelty and redundancy.
    \item Empirical \textbf{evaluation on synthetic and real-world data streams} showing that \ac{FCDC}  increasing data balance by $\SI{25.9}{\percent}$ while reducing data storage by $\SI{39.8}{\percent}$ compared to open-loop collection and random sampling strategies.
\end{enumerate}
Together, these results position \ac{FCDC} as a new data collection paradigm that transforms dataset curation from a passive, one-shot process into an adaptive, self-regulating component of the AI learning loop.

\subsection{Related Work}

Data collection strategies can be classified into data intrinsic and dataset-related strategies. Data intrinsic collection strategies collect data based on its intrinsic properties, which can be based on predefined rules~\cite{elspas_towards_2022},  semantic concepts~\cite{sohn_towards_2025,rigoll_unveiling_2024}  or errors~\cite{gyllenhammar_vehicle_2022}. These strategies do not adapt over time and are static, which work for specific data collection issues, but are not valid for the dynamic process of a balanced dataset collection.

Dataset-related strategies take already collected data into account.
Despite extensive work on active learning~\cite{cacciarelli_active_2024}, coreset selection \cite{guo_deepcore_2022}, and continual learning~\cite{wang_comprehensive_2024}, existing methods address only fragments of this challenge. Active learning focuses on selecting which samples to label, not which to collect and relies on computationally intensive model retraining. Coreset selection methods operate in static batches, which are not compatible with the sequential nature of data streams. Continual learning adapts model parameters rather than managing incoming data itself. 
Another approach for vehicle data collection is anomaly detection, which is systematized in different levels~\cite{20_Breitenstein}. For each level, there exist different approaches for detecting anomalies.
These approaches are either static and require time- and resource-intensive retraining of models or focus on expert-driven anomalies like tiny objects~\cite{xue_novel_2019}, or classes which are not present in a dataset~\cite{chan_segmentmeifyoucan_2021,blum_fishyscapes_2021}.

None of these approaches closes the loop between what has already been collected and what needs to be collected. As a result, current data pipelines lack an intrinsic mechanism to balance exploration of new information with exploitation of existing knowledge. To overcome this limitation, we propose \acf{FCDC}, a paradigm that reframes data collection as a feedback control process. In \ac{FCDC}, the model continuously estimates the density of its observed data and uses this feedback to steer collection decisions. retaining novel samples, down-weighting redundant ones, and adapting collection strategies to maintain a stable level of diversity. This transforms data collection from a static, open-loop activity into an adaptive, self-correcting system capable of autonomously curating high-quality, low-redundancy datasets over time. This FCDC framework generalizes the previous work on adaptive data collection strategies.~\cite{reis_behavior_2024,reis_data-driven_2025}.

\section{Problem Formulation}
Let a data point be $x_i \in \Omega$. Then a continuous data stream from time step $k=0$ to $T$ is denoted by $\mathcal{S}=\left(x[k] \right)_{k=0}^T$, representing sequentially arriving data. A data collection strategy $\mathcal{F}$ maps this stream to a stored dataset $\mathcal{D}$,
\begin{equation*}
    \mathcal{F}:\mathcal{S}_T\rightarrow\mathcal{D}[T]
\end{equation*}
where  $\mathcal{D}_T=\{x[i]\}_{i=1}^N, N\leq T$
 is the subset of data points selected. The objective of the data collection strategy is to ensure that the evolving dataset $\mathcal{D}[k]$  satisfies desired statistical or structural properties $\mathcal{P}$. These properties are quantified by quality metrics
\begin{equation*}
    \mathcal{Q}(\mathcal{D}[k],\mathcal{P})\rightarrow \mathbb{R}
\end{equation*}
which evaluates to what extent the current dataset state exhibits the targeted characteristics, such as high variance in feature embeddings or low redundancy among samples.

The data collection process must be adaptive and dynamic, as both the data distribution of the stream and the dataset composition evolve over time. Consequently, a control policy
\begin{equation*}
    \mathcal{T}:\mathcal{D}[k]\times\Omega\rightarrow\{0,1\}
\end{equation*}
is introduced. The policy $\mathcal{T}$ determines, at each time step, whether an incoming data point $x[k]$ from the data stream $\mathcal{S}$ is collected or rejected from the dataset. 

Due to storage and bandwidth limitations on vehicles, it is infeasible to maintain or repeatedly aggregate the entire dataset $\mathcal{D}[k]$.
Therefore, the system must rely on streaming and incremental estimation of dataset statistics to make efficient, local decisions. Moreover, the decision process operates under the Fast Data paradigm, requiring real-time and iterative updates with minimal computational overhead. Importantly, once a data point is discarded, it cannot be recovered, making each selection decision irreversible. The resulting challenge is to design an online, feedback-driven data collection mechanism that efficiently steers the evolving dataset toward desired properties under these constraints.

\section{Conceptual Framework}
\begin{figure*}[ht]
    \centering
      \includegraphics[width=1\linewidth]{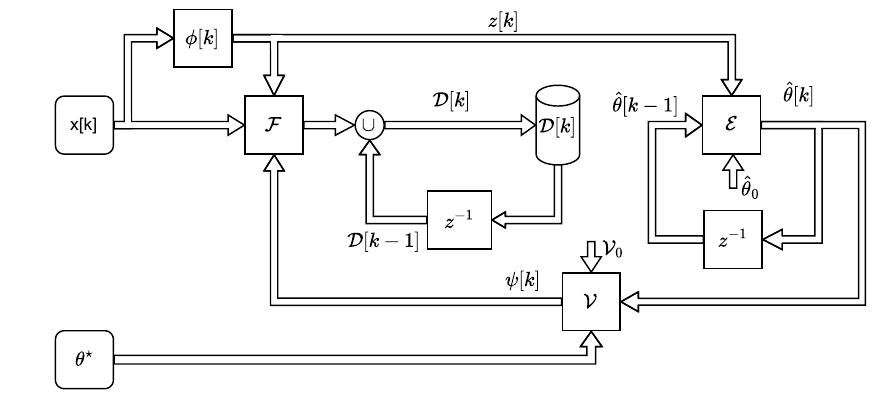}
    \caption{Data Flow Schematic of the Feedback Control Data Collection framework. The input Stream $\mathcal{S}$ consisting of data points $x[k]$ at the corresponding step $k$ as the exogenous disturbance. The Embedding function $\phi$ transforms the data into an embedding space in which the distribution is estimated using the distribution estimation $\mathcal{E}$. Based on the estimation and the target distribution~$\theta^\star$, the Value Function $\mathcal{V}$ updates the data collection strategy in the Collection Control $\mathcal{F}$.}
    \label{fig:fcdc_schematic}
\end{figure*}

The main idea of \ac{FCDC} is to treat the dataset as a dynamical system whose evolution can be controlled through feedback. The incoming data stream is viewed as an exogenous disturbance drawn from a distribution that refers to the Operational Design Domain in the automotive context.  The goal is to steer the dataset toward the desired distribution $\theta^\star$ by adjusting the sampling policy in real-time. Due to the limited vehicle resources, an online dataset estimator maintains a belief over the dataset’s distribution. From this belief, a value function is derived that quantifies the utility of each new observation and defines the collection strategy. Conceptually, \ac{FCDC} functions as a feedback mechanism in which the system estimates and controls the state of its dataset solely based on the feedback provided by the observed data stream (see \autoref{fig:fcdc_schematic}).

\subsection{Components Overview}

\paragraph{Target Distribution $\theta^\star$}
The target distribution $\theta^\star$
 defines the desired statistical state of the dataset. It encodes the dataset properties that the collection process aims to achieve. Conceptually, $\theta^\star$  acts as the reference signal in the feedback loop.

\paragraph{Data Stream $\mathcal{S}$}
Represents the exogenous disturbance as a continuous stream of data points generated by the vehicle’s sensors.
It is an uncontrolled disturbance drawn from the true (and typically unknown) data-generating distribution $p_{true}$ which characterizes the Operational Design Domain (ODD) in the automotive context. This stream defines the pool of potential samples from which the system can decide what to collect.

\paragraph{Dataset $\mathcal{D}$}
This is the state of the system, representing the data that has been collected up to time step $k$. 

\paragraph{Embedding Function $\phi$}
The embedding function $\phi$ transforms a data point into a feature space $\phi(x[k])=z[k]$ where a desired statistic can be evaluated.
This enables the system to compare new observations with the current dataset structure in a desired statistical space. 

\paragraph{Dataset Estimator $\mathcal{E}$}
The estimator maintains an online estimate of the dataset distribution  $\hat\theta\left[k\right]$.
It updates the belief about the dataset state using new incoming embeddings $z[k]$.

\paragraph{Value Function $\psi$}
Based on the estimated dataset distribution $\hat\theta\left[k\right]$ and the target distribution $\theta^\star$, the Value Function $\mathcal{V}$ is computed to update the collection strategy for further incoming data points $x[k]$.
It provides the feedback signal that quantifies how well the current dataset matches the desired distribution.

\paragraph{Collection control $\mathcal{F}$}
This component maps the incoming data representation and the current value function to a \{0,1\} state.
It decides whether a new sample $x[k]$ should be accepted into the dataset based on its evaluated utility.

\subsection{Formal Model}



At each discrete time step $k$, the vehicle observes a new data sample $x[k]$ drawn from the unknown environment distribution $p_{true}$.
The data stream acts as an exogenous disturbance that drives the evolution of the collection process.

The \ac{FCDC} system evolves according to a set of coupled update equations linking the dataset state $\mathcal{D}[k]$, its density estimate $\theta[k]$, and the control policy $\psi[k]$.
These variables interact in closed loop as the current dataset estimate influences the value function, which determines the control action for accepting or rejecting new samples, thereby shaping the future dataset.

The resulting discrete-time nonlinear stochastic dynamics are given by:


\begin{align}
    x[k] &\sim p_{\mathrm{true}} && \text{(exogenous data stream)} \\ 
    z[k] &= \phi(x[k]) && \text{(feature embedding)} \\
    \hat{\theta}[k] &= \mathcal{E}(\hat{\theta}[k-1], z[k]) && \text{(dataset estimation)} \\
    \psi[k] &= \mathcal{V}(\hat{\theta}[k], \theta^\star) && \text{(value function)} \\
    u[k] &= \mathcal{F}(\psi[k], z[k]) && \text{(control input)} \\
    \mathcal{D}[k] &= \mathcal{D}[k-1] \cup u[k] \cdot x[k] && \text{(dataset update)}
\end{align}

Accepted samples (\(u[k] = 1\)) update the dataset state \(\mathcal{D}[k]\) and all incoming data update the estimator parameters \(\hat{\theta}[k]\), which together determine the subsequent value function \(\psi[k]\) and thereby influence future acceptance decisions. 
This closed-loop interaction forms a discrete-time, stochastic control process in which the dataset evolves as the controlled state of the system. 

\section{Implementation Details}

While the embedding function $\phi$ is dependent on the data stream, the dataset estimation $\mathcal{E}$, value function $\mathcal{V}$, and the control Function $\mathcal{F}$ remain fixed for the experiments. 
\subsection{State Estimation}
 The distribution of the incoming stream is modeled online by a d-dimensional Gaussian
\begin{equation}\label{eq:multi_gaussian}
    \mathcal{N}(z,\mu,\Sigma) = \underbrace{\frac{1}{2\pi^{d/2}\vert\Sigma\vert^{1/2}}}_{\beta}\exp{\left(-\frac{1}{2} (z-\mu)^\top\Sigma^{-1}(z-\mu) \right)} 
\end{equation}
with parameters $\mu[k]$ and $\Sigma[k]$ updated incrementally from each observed sample using Welford's algorithm~\cite{Welford1962NoteOA}:

\begin{align}
    n[k] &\leftarrow n[k-1] + 1, \\
    \delta[k] &\leftarrow x[k] - \mu[k-1], \\
    \mu[k] &\leftarrow \mu[k-1] + \frac{\delta[k]}{n[k]},\label{eq:est_meean} \\
    C[k] &\leftarrow C[k-1] + \delta[k] \cdot\left(x[k] - \mu[k]\right)^\top\\
    {\Sigma}[k]& \leftarrow \frac{C[k]}{n[k] - 1}, \quad \text{for } n[k] > 1. \label{eq:est_covariance}
\end{align}

\subsection{Value Function}
To showcase the controllability of the \ac{FCDC}, two different value functions are used:\\

\subsubsection{Complementary Gaussian $\psi_\mathrm{C}$} 
The complementary value function is given by $\psi_\mathrm{C}= 1-\mathcal{N}(\cdot)$ normalized by the maximum value and adjusted with a warm-up factor 
$\kappa$ to ensure stability when the sample size is small $N$. Given the  Mahalanobis Distance $d_\mathrm{M}$
\begin{equation} \label{eq:d_M}
    d_\mathrm{M}(\mu,z)=\sqrt{(z-\hat{\mu})^\top{\Sigma}^{-1}(z-\hat{\mu})},
\end{equation} 
the value function can be solved in closed form as
\begin{align}
    \psi_\mathrm{C}(z) &= 1-\underbrace{\min\left(1,\frac{\vert N\vert}{\nu}\right)}_{\kappa} \cdot\left(\frac{\mathcal{N}(\hat{\theta},z)}{\max\left(\mathcal{N}(\hat{\theta })\right)}\right) \nonumber \\
    &=1-\kappa\cdot\exp{\left(-\frac{1}{2}d_\mathrm{M}(\mu,z)^2\right)},\label{eq:val_fcn}
\end{align}

\begin{proof}
    Given  $d$-dimensional Gaussian from \eqref{eq:multi_gaussian}, then the ratio between the distribution at a specific point and its $\max$ value located at the mean is given by 
    \begin{equation*}
    \begin{split}
        \frac{\mathcal{N}(\hat{\theta},z)}{\max\left(\mathcal{N}(\hat{\theta })\right)} &=
        \frac{\beta\cdot\exp{\left(-\frac{1}{2} (z-\mu)^\top\Sigma^{-1}(z-\mu) \right)} }{\beta\cdot\exp(0)} \\
        &=\exp{\left(-\frac{1}{2} (z-\mu)^\top\Sigma^{-1}(z-\mu) \right)} \\
        &\overset{\eqref{eq:d_M}}{=} \exp{\left(-\frac{1}{2}d_\mathrm{M}(\mu,z)^2\right)},
    \end{split}
    \end{equation*}
    which leads to \eqref{eq:val_fcn}.
\end{proof}
Since the covariance $\Sigma$ in~\eqref{eq:d_M} may be ill-conditioned or near-singular (especially in higher order dimensionality), a covariance estimation using Ledoit-Wolf  shrinkage~\cite{ledoit_honey_2003}  may be necessary, as demonstrated in \cite{reis_data-driven_2025}. 

\subsubsection{Reciprocal Gaussian $\psi_\mathrm{R}$}

To represent a uniform target distribution over a bounded region $D \subset \mathbb{R}^2$, we define $D$ as a Mahalanobis ellipse,
\begin{equation*}
D = \{\, z \mid d_M^2(z) \le r_{\max}^2 \,\},
\end{equation*}
where $d_M^2(z)$ denotes the Mahalanobis distance and $r_{\max}^2$ controls the spatial extent of $D$.
Given the Gaussian estimation from \eqref{eq:multi_gaussian} and a uniform target $\theta^\star(z) = c_D$ within $D$, the corresponding value function is derived from the ratio between the target and estimated distributions,
\begin{equation*}
\psi(z) \propto \frac{\theta^\star(z)}{q(z)} \propto \exp\!\left(\tfrac{1}{2} d_M^2(z)\right).
\end{equation*}
After normalization over $D$, this yields the final form
\begin{equation}
\label{eq:uniform_value_function}
\psi_\mathrm{R}(z) =
\begin{cases}
\exp\!\left(\tfrac{1}{2}\bigl(d_M^2(z) - r_{\max}^2\bigr)\right), & \text{if } d_M^2(z) \le r_{\max}^2, \\[0.8ex]
0, & \text{otherwise.}
\end{cases}
\end{equation}

\subsection{Control Function}
Based on the Value Function $\psi$ from \eqref{eq:val_fcn} and \eqref{eq:uniform_value_function}, the control function $u[k]$ determines whether a sample is collected. 
The sampling process is modeled as a Bernoulli trial with success probability $\psi$:

\begin{equation}
    u[k] \sim \mathrm{Bernoulli}(\psi(z[k])),
    \quad 
    P(u[k] = 1) = \psi[k],
    \label{eq:control_function}
\end{equation}
where $u[k] = 1$ indicates that the sample is collected and $u[k] = 0$ otherwise.
This stochastic decision is implemented by drawing 
$r \sim \mathcal{U}(0,1)$ and setting $u[k] = 1$ if $r < \psi(z[k])$, otherwise $u[k] = 0$.

\section{Evaluation}
\begin{figure*}[t]
    \centering
    \begin{subfigure}[b]{1\textwidth}
    \includegraphics[width=1\linewidth]{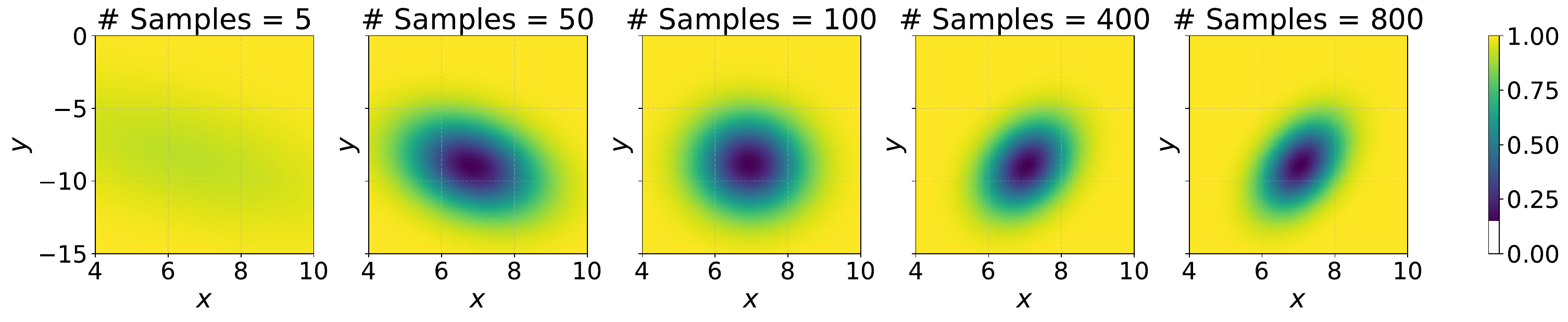}
    \label{fig:value_fcn_density}
    \end{subfigure}
    \vfill
    \centering
    \begin{subfigure}[b]{1\textwidth}
    \includegraphics[width=1\linewidth]{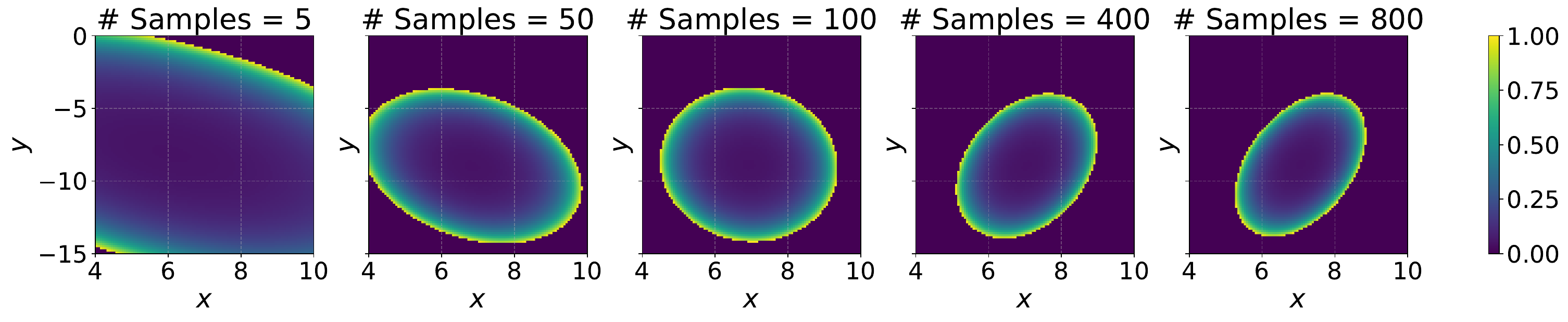}
    \label{fig:value_fcn_uniform}
    \end{subfigure}
    \caption{Evolution of the collection probability over the number of processed samples for newly arriving data points, based on $\psi_\mathrm{C}$ (top) and  $\psi_U$ (bottom).}
    \label{fig:value_functions}
\end{figure*}
\begin{figure*}[h!]
    \centering
    \begin{subfigure}[b]{0.3\textwidth}
        \includegraphics[width=\textwidth]{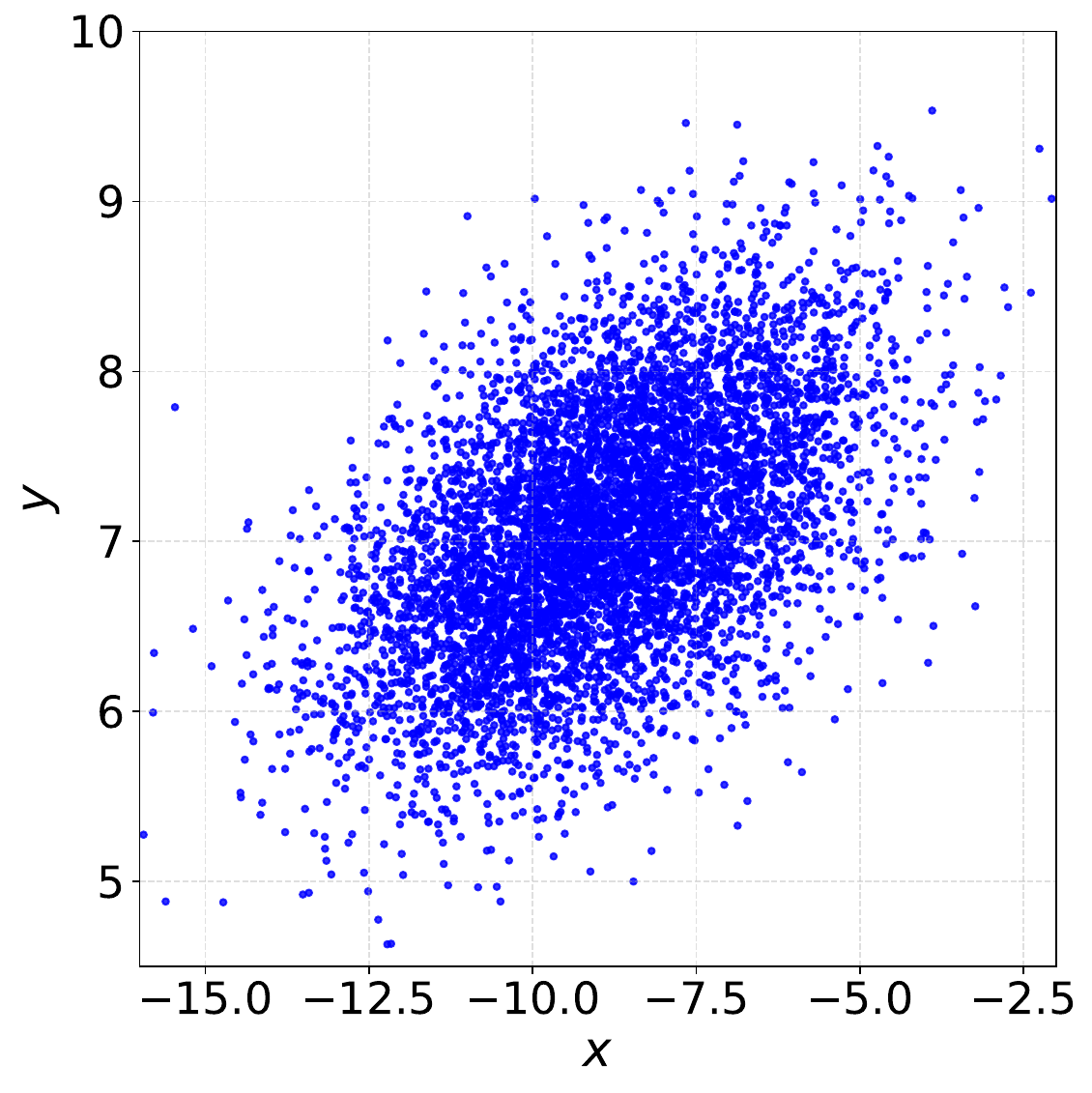}
        \label{fig:1a}
    \end{subfigure}
    \hfill
    \begin{subfigure}[b]{0.3\textwidth}
        \includegraphics[width=\textwidth]{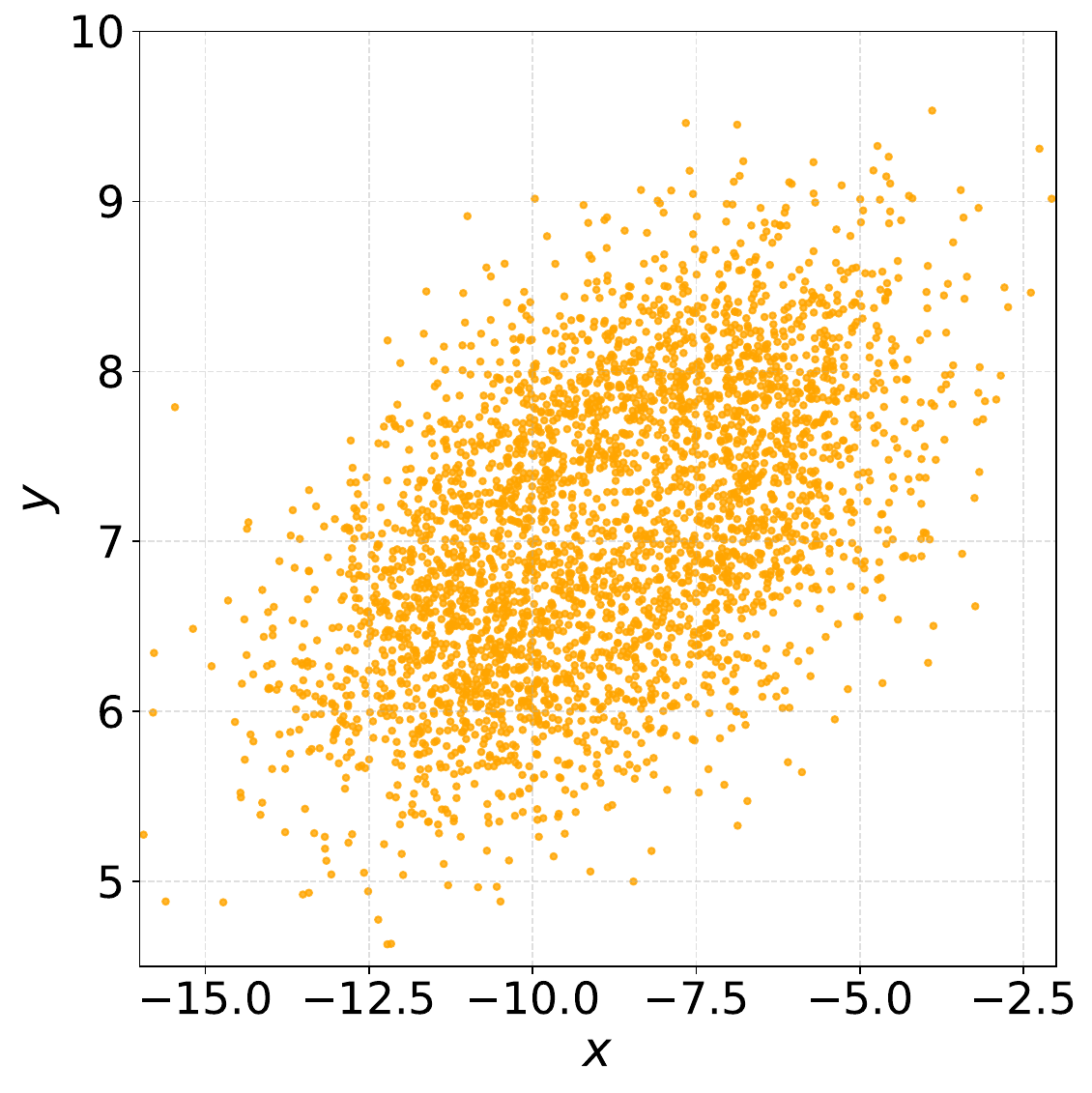}
        \label{fig:1b}
    \end{subfigure}
    \hfill
    \begin{subfigure}[b]{0.3\textwidth}
        \includegraphics[width=\textwidth]{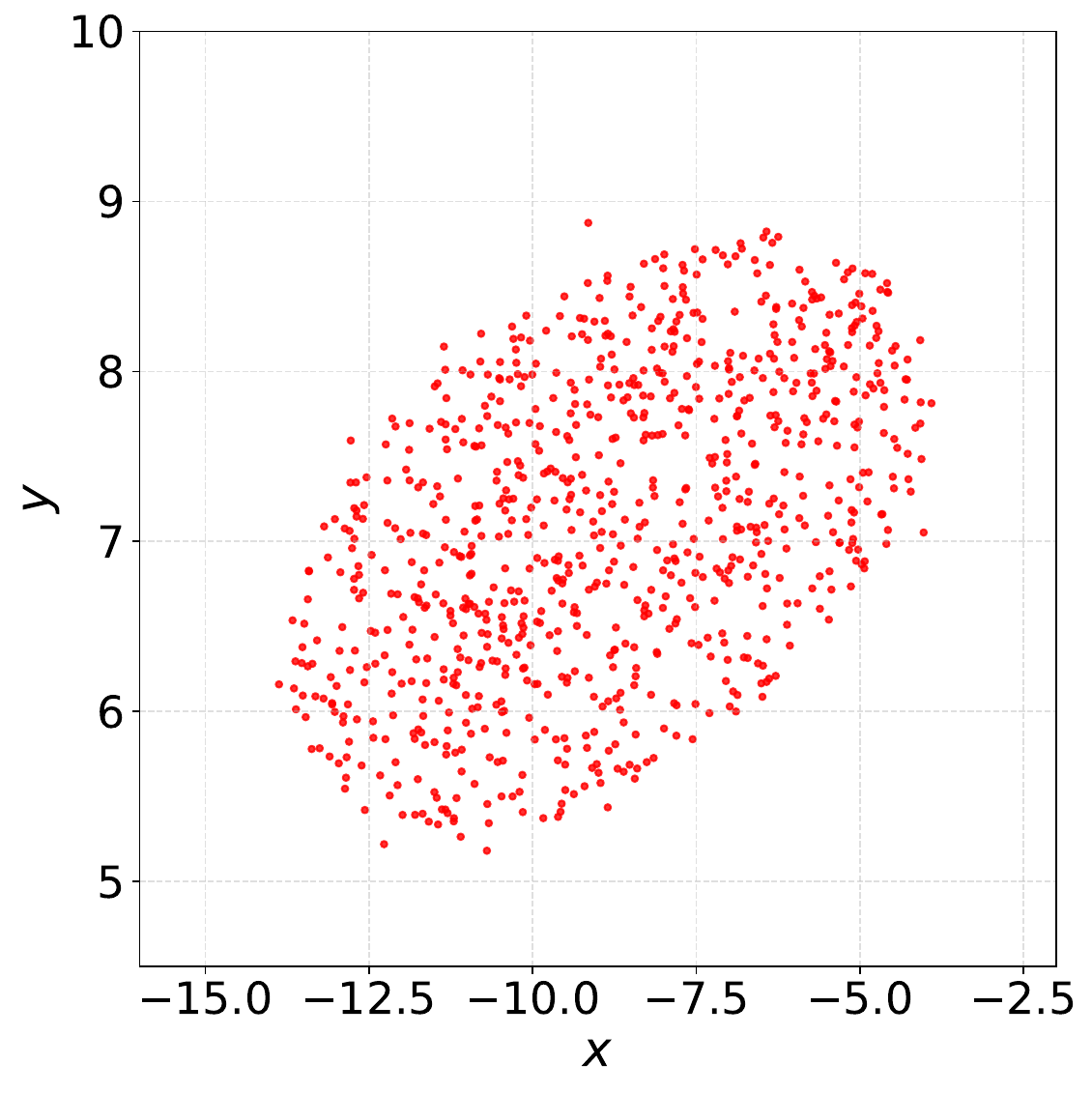}
        \label{fig:1c}
    \end{subfigure}
    \caption{Comparison of selected data points between the ground truth (\mytikzDotBlue), $\psi_\mathrm{C}$  (\mytikzDotOrange) and $\psi_\mathrm{R}$ (\mytikzDotRed).}
    \label{fig:scatter_comparison}
\end{figure*}

The evaluation consists of two experiments, one using synthetic 2D data with two distinct value functions for showcasing the controllability of the target dataset distribution using a multivariate Gaussian as ground truth density, and the second experiment is conducted on the real vehicle system CoCar nextGen~\cite{Heinrich_24}, which targets a balanced collection of the number of vehicles per image.

\subsection{Synthetic Datastream Experiment}
\begin{figure}[t]
    \centering
    \includegraphics[width=1\linewidth]{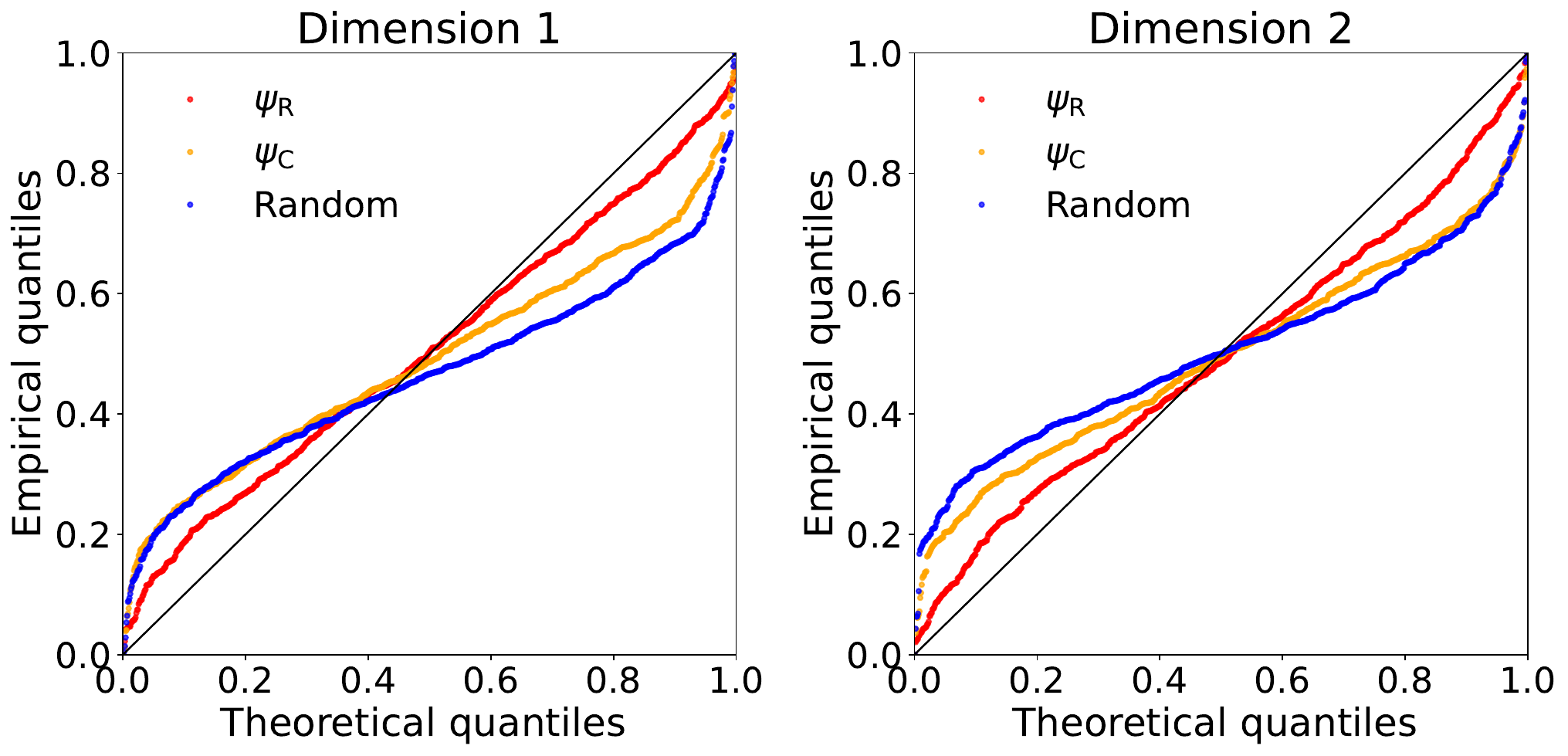}
    \caption{Q–Q diagrams comparing $10^5$ samples: random collection (\mytikzDotBlue), FCDC with $\psi_\mathrm{C}$ (\mytikzDotOrange), and FCDC with $\psi_\mathrm{R}$ (\mytikzDotRed), each compared against a theoretical uniform distribution in both dimensions. Deviation of the diagonal means a deviation to the theoretical uniform distribution.}
    \label{fig:Q-Q_diagram}
\end{figure}

The synthetic data stream consists of 6,000 samples drawn from a two-dimensional Gaussian distribution, which serves as the ground truth (see \autoref{fig:scatter_comparison}, left). Since the data are already defined in the target space, the embedding function $\phi$ is set to the identity map. The samples are sequentially processed by the \ac{FCDC} framework.
For the uniform target distribution with $\psi_\mathrm{R}$, a total of 877 points are collected, whereas for $\psi_\mathrm{C}$, 3,353 points are retained (see \autoref{fig:scatter_comparison}, middle and right). Using $\psi_\mathrm{R}$, the resulting data distribution approximates the desired uniform target distribution. However, data points located outside the collection region, defined by $d_{M}^2(z) > r^2_{\max}$, are excluded, potentially omitting informative samples. In contrast, case $\psi_\mathrm{C}$ reduces the number of samples in high-density regions near the estimated mean, but the overall collected data are not uniformly distributed.
The experiments can be separated into two phases: during the convergence of the data distribution estimation and after convergence is achieved. As the number of processed samples increases, the value function also converges, reflecting the improved estimation of the underlying dataset distribution (see \autoref{fig:value_functions}, with  $\psi_\mathrm{C}$ shown at the top and $\psi_\mathrm{R}$ at the bottom). Although the dataset distribution estimation $\mathcal{E}$ remains consistent across both value functions $\psi[k]$, their influence on the data collection process and resulting data distributions differ, as illustrated in~\autoref{fig:Q-Q_diagram}. 
Based on the conformity to a uniform distribution, the root-mean-square error $r_j$ for each dimension is computed between the sorted empirical quantiles $m_i$ and the theoretical quantiles $p_i$, resulting in the total error $\Delta_\mathrm{uni}$ as
\begin{equation}
      \Delta_\mathrm{uni}=\sqrt{\frac{1}{d}\sum_{j=1}^dr_j^2}, \quad r_j = \sqrt{\frac{1}{n}\sum_{i=1}^n(u_i-p_i)^2},. 
\end{equation}

\begin{figure}[t]
    \centering
    \includegraphics[width=1\linewidth]{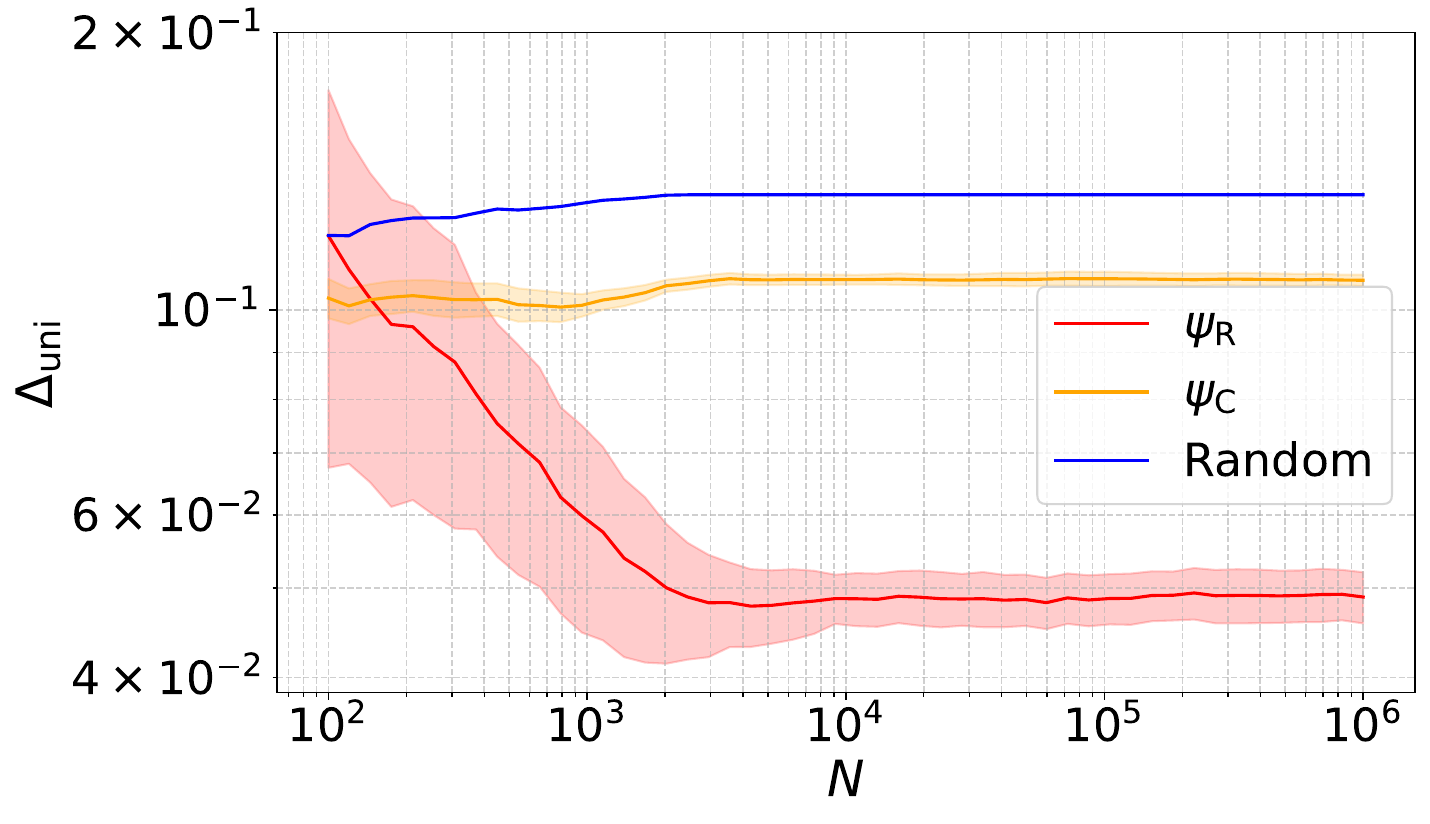}
    \caption{Error $\Delta_\mathrm{uni}$ of collected samples $N$ to a theoretical uniform over the number of collected samples using random collection (\mytikzDotBlue), FCDC with $\psi_\mathrm{C}$ (\mytikzDotOrange), and FCDC with $\psi_\mathrm{R}$ (\mytikzDotRed).}
    \label{fig:error}
\end{figure}

The error converges for $\psi_\mathrm{R}$ as the number of samples increases, whereas $\psi_\mathrm{C}$ stays at a constant value but below the random sampling, indicating improvement and redundancy reduction, see~\autoref{fig:error}.

\begin{figure*}[t]
    \centering
    \includegraphics[width=0.93\linewidth]{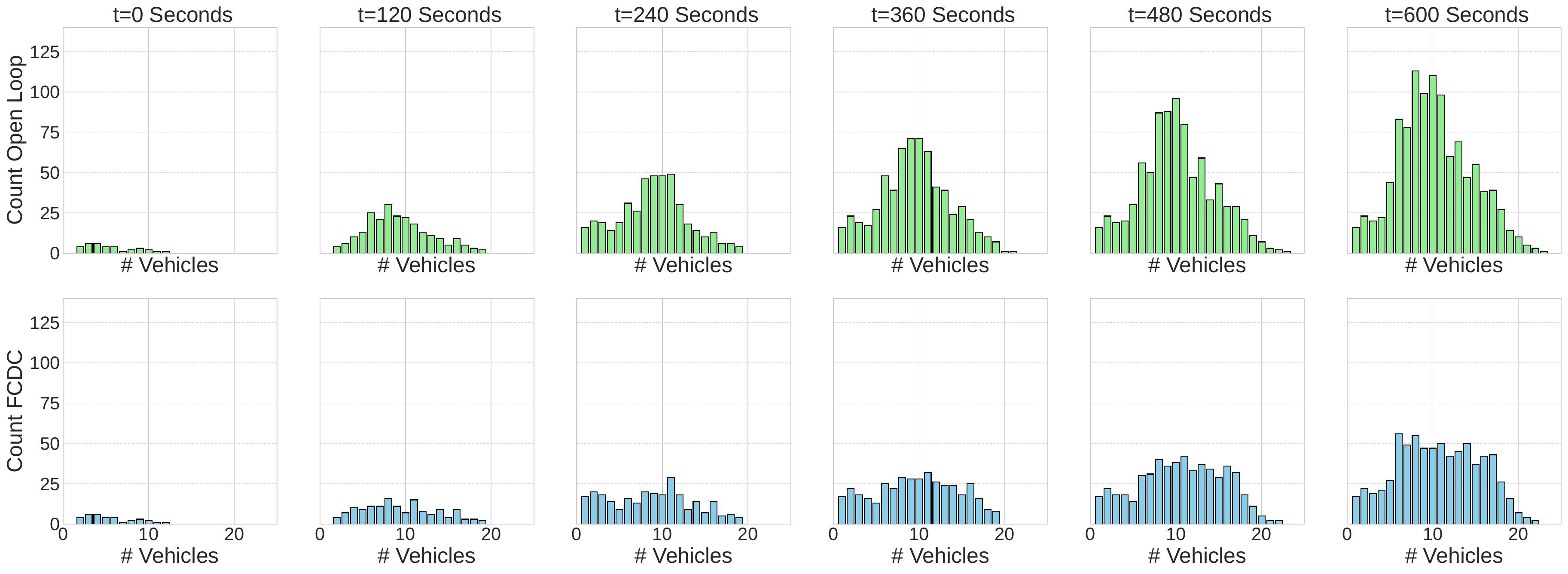}
    \caption{Evolution of the number of vehicles per image for open-loop collection (top) compared with the FCDC strategy (bottom). }
    \label{fig:count_histo}
\end{figure*}

\begin{figure}[t]
    \centering
    \includegraphics[width=0.8\linewidth]{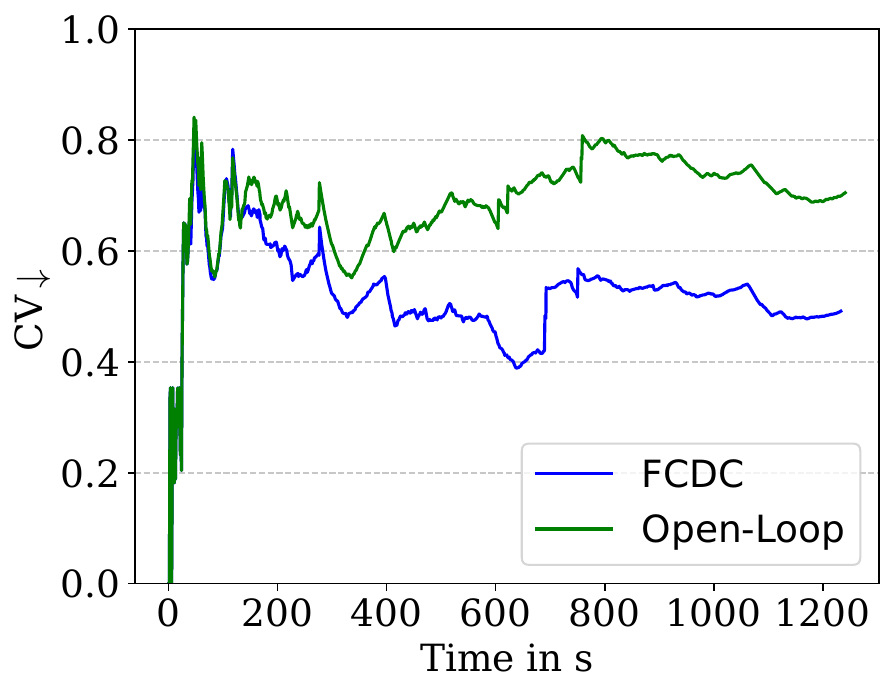}
    \caption{Coefficient of variation over time of FCDC compared to the open-loop collection strategy. Lower values indicate a more balanced dataset.}
    \label{fig:cv_over_time}
\end{figure}
In both feedback strategies, the deviation from a uniform distribution is smaller than in the random collection strategy. Notably, the deviation of the complementary Gaussian $\psi_\mathrm{C}$ is more pronounced compared to the Reciprocal Gaussian  $\psi_\mathrm{R}$. As a result, the Reciprocal value function yields more uniform data distribution during the collection process. However, due to the underlying value function, out-of-distribution samples are discarded, which may represent valuable data candidates. In contrast, $\psi_\mathrm{C}$ collects data across the entire latent space at the cost of a higher deviation from a uniform distribution.

\subsection{Real-Data Experiment}

For the real-world dataset, a 20-minute data stream from the CoCar NextGen platform \cite{Heinrich_24} was analyzed, with a total of 1356 images. The objective of this experiment is to construct an image dataset that is balanced with respect to the number of vehicles per image, covering the discrete range $c \in \{0, \ldots, K\}$ visible vehicles per image.  
The embedding function $\phi$ comprises a sequential processing pipeline: first, a YOLO-based object detection is applied to each incoming image; second, a counting function $count(\cdot)$ determines $z$, the number of detected vehicles. The corresponding image should be collected if it contributes to the overall uniform data distribution. 
 
The \ac{FCDC} employs the value function defined in \eqref{eq:val_fcn}. A comparison between the data distributions obtained through the \ac{FCDC} and an open-loop data collection strategy is presented in \autoref{fig:count_histo}. While the open-loop approach overrepresents samples containing 8-11 vehicles per image, the \ac{FCDC} framework compensates for this overrepresentation through its feedback mechanism, thereby reducing sampling in this region. In both cases, images containing more than 17 vehicles per frame remain underrepresented, as such scenes are inherently rare.

The balance of the dataset throughout the collection process is monitored using the Coefficient of Variation (CV), defined as $CV(\mathcal{D})= {\mu}/{\sigma}$, see \autoref{tab:comparison_FCDC_OL}. Initially, the balance is the same due to the warm-up phase of the feedback control. Over time, the divergence of CV indicates that the \ac{FCDC} strategy consistently produces a more balanced dataset compared to the open-loop approach. Overall, the \ac{FCDC} method collected $\SI{39.8}{\percent}$ fewer samples while improving the dataset balance, as measured by the coefficient of variation, by $\SI{25.9}{\percent}$.

\begin{table}[t]
    \centering
    \caption{Comparison of the open-loop data collection and the FCDC strategy.}
    \label{tab:comparison_FCDC_OL}
    \begin{tabularx}{0.8\linewidth}{l c c} 
        \toprule
        Metric & Open-loop & FCDC \\
        \midrule
        \# Data points & 1356 & 817 \\
        Coefficient of Variation $\downarrow$ & 0.73 & 0.541 \\
        \bottomrule
    \end{tabularx}
\end{table}

\section{Conclusion and Outlook}\label{sec:Conclusion}
This work introduced the Feedback Control Data Collection (FCDC) framework, which formulates data collection as a dynamic system whose state evolves under the influence of an incoming data stream treated as an exogenous disturbance. Through continuous estimation of the current dataset distribution and comparison with a target distribution, the framework derives a feedback signal that adapts the collection strategy in real-time.

Experimental results demonstrate that feedback control can steer the distribution of collected data toward a desired target. Two value functions are investigated: a reciprocal and a complementary Gaussian. Both reduced redundancy and improved convergence toward a uniform target distribution. The reciprocal value function enforces stricter control, converging to a uniform but potentially less diverse dataset. In contrast, the complementary Gaussian function provides a more flexible response, achieving closer alignment to the uniform distribution than random sampling yet allowing greater variability than the reciprocal function.
Future research should focus on extending FCDC to multidimensional and multimodal data distributions, establishing end-to-end evaluations that link value function design to model performance. Together, these directions will advance FCDC toward a framework for autonomous, self-regulating data collection in intelligent vehicle systems.
\section{ACKNOWLEDGMENT}
This work results from the just better DATA (jbDATA)
project supported by the German Federal Ministry for Economic Affairs and Climate Action of Germany (BMWK) and the European Union, grant number 19A23003H.
\bibliography{references.bib}
\end{document}